\documentclass[graybox]{svmult} 

\usepackage{type1cm} 
\usepackage{makeidx} 
\usepackage{graphicx} 
\usepackage{multicol} 
\usepackage[bottom]{footmisc} 
\usepackage{amsmath,amssymb}
\usepackage{bm}
\usepackage[numbers]{natbib}
\usepackage{url}
\usepackage{tikz}

\usepackage[framemethod=TikZ]{mdframed}
\definecolor{graybox}{rgb}{.9,.9,.9}
\mdfsetup{roundcorner=4,linewidth=0,backgroundcolor=graybox,leftmargin=0em,rightmargin=0em,
innerleftmargin=0.3em,innerrightmargin=0.3em,innertopmargin=0.3em,innerbottommargin=0.3em}

\makeindex 

\newcommand{\trsp}{{\scriptscriptstyle\top}}
\newcommand{\psin}{{\dagger}}

\newcommand{\ty}[1]{{\scriptscriptstyle{#1}}}
\newcommand{\diag}{\mathrm{diag}}
\newcommand{\tr}{\mathrm{tr}}

\begin{document}

\title*{Mixture models for the analysis, edition, and synthesis of continuous time series}
\titlerunning{Mixture models for continuous time series}
\author{Sylvain Calinon}
\institute{Sylvain Calinon \at Idiap Research Institute, Martigny, Switzerland, \email{sylvain.calinon@idiap.ch}}
\maketitle

\begin{tikzpicture}[remember picture,overlay]
\node[text width=140mm, inner sep=40, anchor=south west] at (current page.south west) 
{\emph{This is the author's version of ``Mixture models for the analysis, edition, and synthesis of continuous time series''. The final publication is available at Springer in the book ``Mixture Models and Applications'' (2019), pp. 39--57, edited by Bouguila, N. and Fan, W.}};
\end{tikzpicture}

\abstract{This chapter presents an overview of techniques used for the analysis, edition, and synthesis of continuous time series, with a particular emphasis on motion data. The use of mixture models allows the decomposition of time signals as a superposition of basis functions. It provides a compact representation that aims at keeping the essential characteristics of the signals. Various types of basis functions have been proposed, with developments originating from different fields of research, including computer graphics, human motion science, robotics, control, and neuroscience. Examples of applications with radial, Bernstein and Fourier basis functions are presented, with associated source codes to get familiar with these techniques.}

\section{Introduction}
\label{sec:intro}
The development of techniques to process continuous time series is required in various domains of application, including computer graphics, human motion science, robotics, control, and neuroscience. These techniques need to cover various purposes, including the encoding, modeling, analysis, edition, and synthesis of time series (sometimes needed simultaneously). The development of these techniques is also often governed by additional important constraints such as interpretability and reproducibility. These heavy requirements motivate the use of mixture models, effectively leveraging the formalism and ubiquity of these models. 

The first part of this chapter reviews decomposition techniques based on radial basis functions (RBFs) and locally weighted regression (LWR). The connections between LWR and Gaussian mixture regression (GMR) are discussed, based on the encoding of time series as Gaussian mixture models (GMMs). I will show how this mixture modeling principle can be extended to a weighted superposition of Bernstein basis functions, often known as B\'ezier curves. The aim is to examine the connections with mixture models and to highlight the generative aspects of these techniques. In particular, this link exposes the possibility of representing B\'ezier curves with higher order Bernstein polynomials. 
I then discuss the decomposition of time signals as Fourier basis functions, by showing how a mixture of Gaussians can leverage the multivariate Gaussian properties in the spatial and frequency domains. Finally, I show that these different decomposition techniques can be represented as time series distributions through a probabilistic movement primitives representation. 

Pointers to various practical applications are provided for further readings, including the analysis of biological signals in the form of multivariate continuous time series, the development of computer graphics interfaces to edit trajectories and motion paths for manufacturing robots, the analysis and synthesis of periodic human gait data, or the generation of exploratory movements in mobile platforms with ergodic control.

The techniques presented in this chapter are described with a uniform notation that does not necessarily follow the original notation. The goal is to tie links between these different techniques, which are often presented in isolation of the more general context of mixture models. Matlab codes accompany the chapter~\cite{pbdlib}, with full compatibility with GNU Octave.

\section{Movement primitives}
\label{sec:primitives}
\begin{figure}
  \centering
  \includegraphics[width=.8\columnwidth]{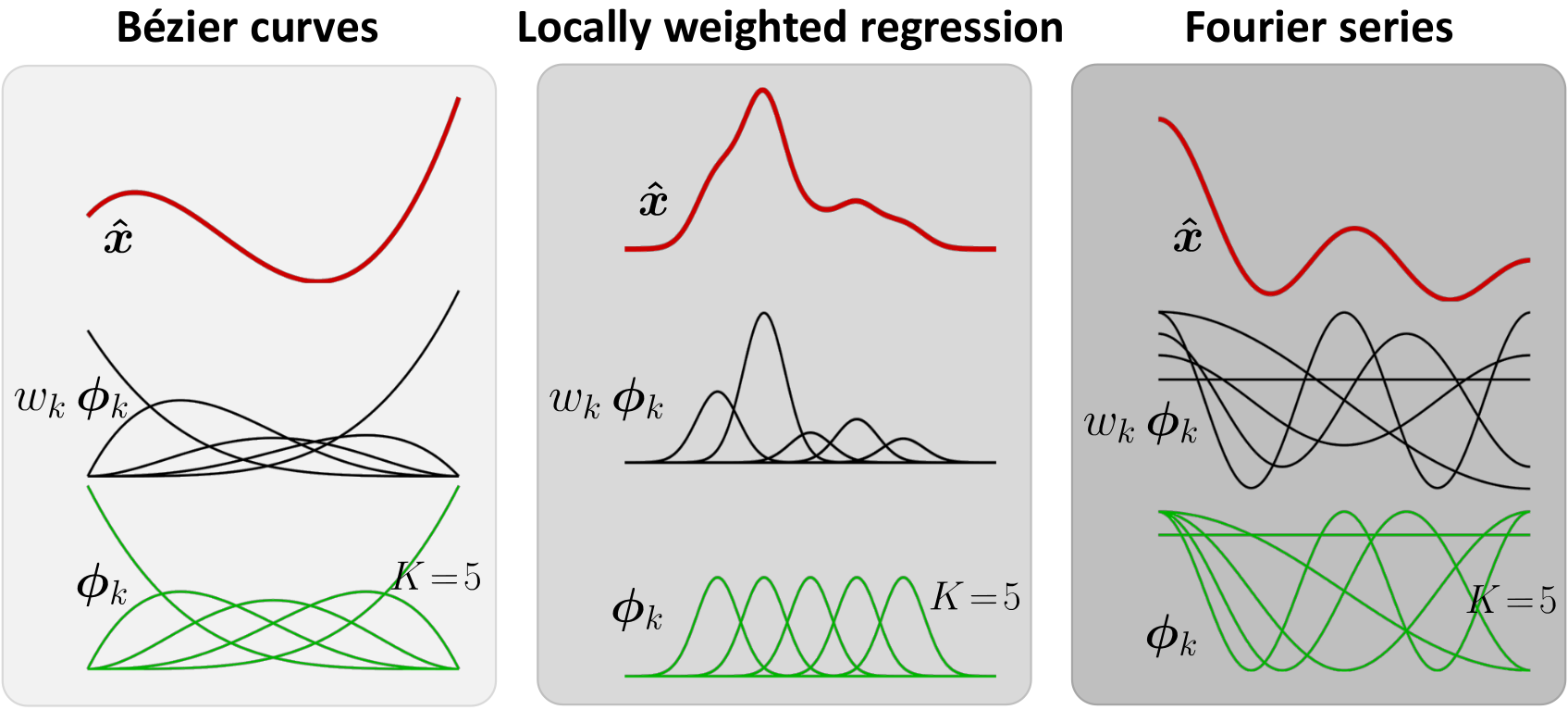}
  \caption{
Motion primitives with different basis functions $\phi_k$, where a unidimensional time series $\bm{\hat{x}}=\sum_{k=1}^K w_k \phi_k$ is constructed as a weighted superposition of $K$ signals $\phi_k$.
  }
  \label{fig:superposition}
\end{figure}

The term \emph{movement primitives} refers to an organization of continuous motion signals in the form of a superposition in parallel and in series of simpler signals, which can be viewed as ``building blocks'' to create more complex movements, see Fig.~\ref{fig:superposition}. This principle, coined in the context of motor control~\cite{MussaIvaldi94}, remains valid for a wide range of continuous time signals (for both analysis and synthesis). Next, I present three popular families of basis functions that can be employed for time series decomposition.
\subsection{Radial basis functions (RBFs)}
\label{sec:RBF}

\begin{figure}
  \centering
  \includegraphics[width=.8\columnwidth]{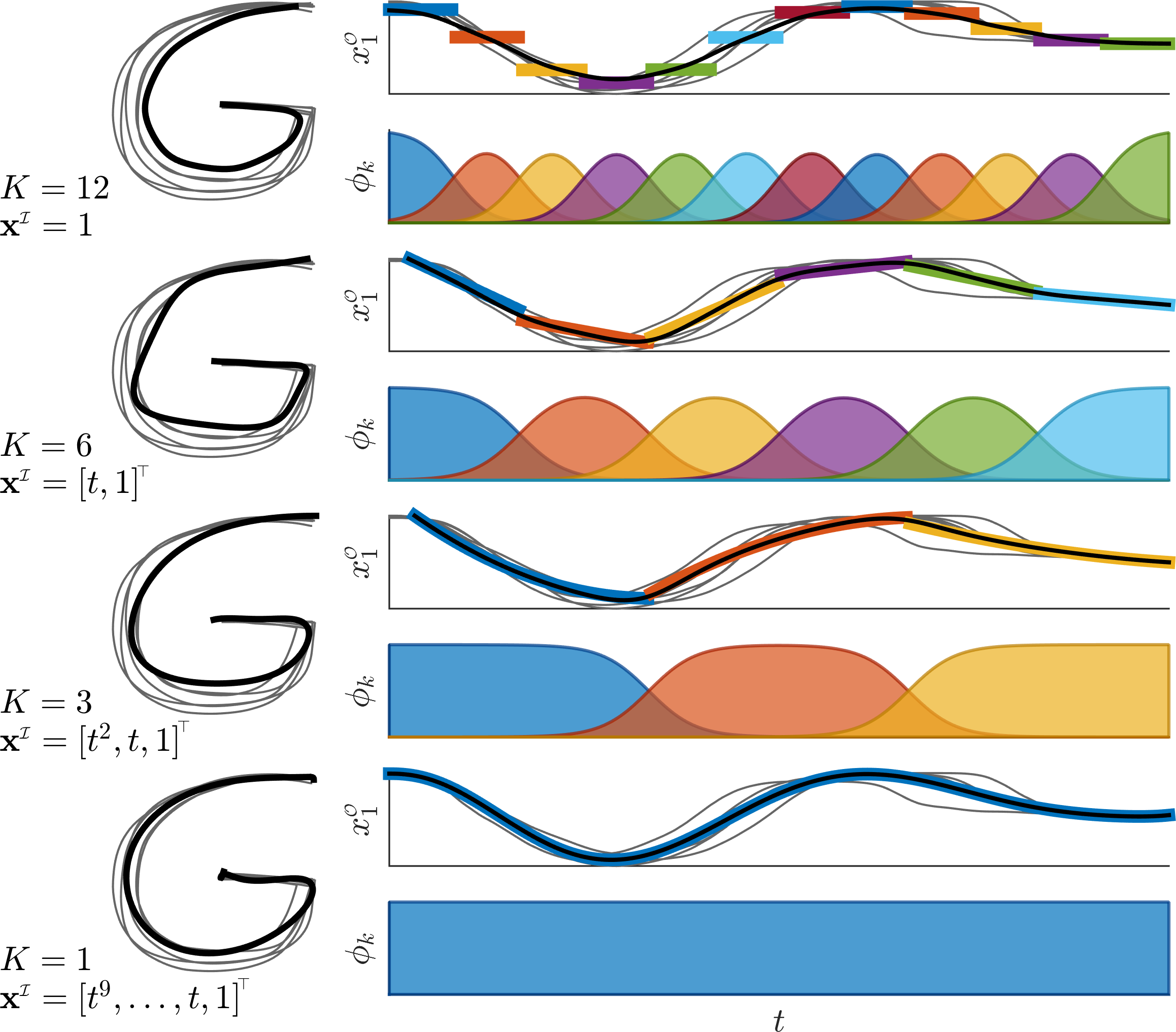}
  \caption{
Polynomial fitting with locally weighted regression (LWR), by considering different degrees of the polynomial and by adapting the number of basis functions accordingly. The top row shows a very localized encoding of the movement, with constant values used in Eq.~\eqref{eq:WLS1}, thus requiring the use of many basis functions to represent the trajectory. In the top timeline graph, the colored horizontal segments can also be interpreted as scalar weights used to approximate the original signal by a weighted superposition of radial basis functions (see example in Fig.~\ref{fig:superposition}).
The next rows show that a reduction of this number of basis functions typically needs to be compensated with more complex basis functions (polynomial of higher degrees). The bottom row depicts the limit case in which a global encoding of the movement would require a polynomial of high degree.
  }
  \label{fig:LWR}
\end{figure}

\noindent Radial basis functions (RBFs) are ubiquitous in continuous time series encoding~\cite{Stulp15}, notably due to their simplicity and ease of implementation. Most algorithms exploiting this representation rely on some form of regression, often related to locally weighted regression (LWR), which was introduced by~\cite{Cleveland79} in statistics and popularized by~\cite{Atkeson90} in robotics. By representing, respectively, $N$ input and output datapoints as $\bm{X}^\ty{I}={[\bm{x}_1^\ty{I},\bm{x}_2^\ty{I},\ldots,\bm{x}_N^\ty{I}]}^\trsp$ and $\bm{X}^\ty{O}={[\bm{x}_1^\ty{O},\bm{x}_2^\ty{O},\ldots,\bm{x}_N^\ty{O}]}^\trsp$, we are interested in the problem of finding a matrix $\bm{A}$ so that $\bm{X}^\ty{I}\bm{A}$ would match $\bm{X}^\ty{O}$ by considering different weights on the input--output datapoints $\{\bm{X}^\ty{I},\bm{X}^\ty{O}\}$ (namely some datapoints are more informative than others for the estimation of $\bm{A}$). A weighted least squares estimate $\bm{\hat{A}}$ can be found by solving the objective
\begin{align}
	\bm{\hat{A}} &= \arg\underset{\bm{A}}{\min}\; 
	\tr\Big({(\bm{X}^\ty{O}-\bm{X}^\ty{I}\bm{A})}^\trsp \bm{W} (\bm{X}^\ty{O}-\bm{X}^\ty{I}\bm{A})\Big)\nonumber\\ 
	&= {({\bm{X}^\ty{I}}^{\trsp}\bm{W}\bm{X}^\ty{I})}^{-1}{\bm{X}^\ty{I}}^{\trsp}\bm{W} \,\bm{X}^\ty{O},
	\label{eq:WLS1}
\end{align} 
where $\bm{W}\!\in\!\mathbb{R}^{N\!\times\!N}$ is a weighting matrix. Locally weighted regression (LWR) is a direct extension of the weighted least squares formulation in which $K$ weighted regressions are performed on the same dataset $\{\bm{X}^\ty{I},\bm{X}^\ty{O}\}$. It aims at splitting a nonlinear problem so that it can be solved locally by linear regression. LWR computes $K$ estimates $\bm{\hat{A}}_k$, each with a different function $\phi_k(\bm{x}^\ty{I}_n)$, classically defined as the radial basis functions
\begin{equation}
	\tilde{\phi}_k(\bm{x}^\ty{I}_n) = \exp\Big(\!-\frac{1}{2} {(\bm{x}^\ty{I}_n-\bm{\mu}^\ty{I}_k)}^\trsp 
	{\bm{\Sigma}^\ty{I}_k}^{-1} (\bm{x}^\ty{I}_n-\bm{\mu}^\ty{I}_k) \Big),
	\label{eq:rbf}
\end{equation}
where $\bm{\mu}^\ty{I}_k$ and $\bm{\Sigma}^\ty{I}_k$ are the parameters of the $k$-th RBF, or in its rescaled form\footnote{We will see later that the rescaled form is required for some techniques, but for locally weighted regression, it can be omitted to enforce the independence of the local function approximators.}   
\begin{equation}
	\phi_k(\bm{x}^\ty{I}_n) = \frac{\tilde{\phi}_k(\bm{x}^\ty{I}_n)}
	{\sum_{i=1}^K \tilde{\phi}_i(\bm{x}^\ty{I}_n)}.
	\label{eq:rbf2}
\end{equation}
An associated diagonal matrix
\begin{equation}
	\bm{W}_k = \mathrm{diag}\Big(\phi_k(\bm{x}^\ty{I}_1),\phi_k(\bm{x}^\ty{I}_2),\ldots,\phi_k(\bm{x}^\ty{I}_N)\Big)
\end{equation}
can be used with \eqref{eq:WLS1} to evaluate $\bm{\hat{A}}_k$. The result can then be employed to compute 
\begin{equation}
	\bm{\hat{X}}^\ty{O} = \sum_{k=1}^K \bm{W}_k \, \bm{X}^\ty{I} \bm{\hat{A}}_k.
\end{equation}
The centroids $\bm{\mu}^\ty{I}_k$ in \eqref{eq:rbf} are usually set to uniformly cover the input space, and $\bm{\Sigma}^\ty{I}_k\!=\!\bm{I}\sigma^2$ is used as a common bandwidth shared by all basis functions. Figure \ref{fig:LWR} shows an example of LWR to encode planar trajectories.

LWR can be directly extended to local least squares polynomial fitting by changing the definition of the inputs. Multiple variants of the above formulation exist, including online estimation with a recursive formulation~\cite{Schaal98}, Bayesian treatments of LWR~\cite{Ting08}, or extensions such as locally weighted projection regression (LWPR) that exploit partial least squares to cope with redundant or irrelevant inputs~\cite{Vijayakumar05}. 

Examples of application range from inverse dynamics modeling~\cite{Vijayakumar05} to the skillful control of a devil-stick juggling robot~\cite{Atkeson97}.
A Matlab code example \texttt{demo\_LWR01.m} can be found in~\cite{pbdlib}.

\subsubsection{Gaussian mixture regression (GMR)} 

\begin{figure}
  \centering
  \includegraphics[width=.65\columnwidth]{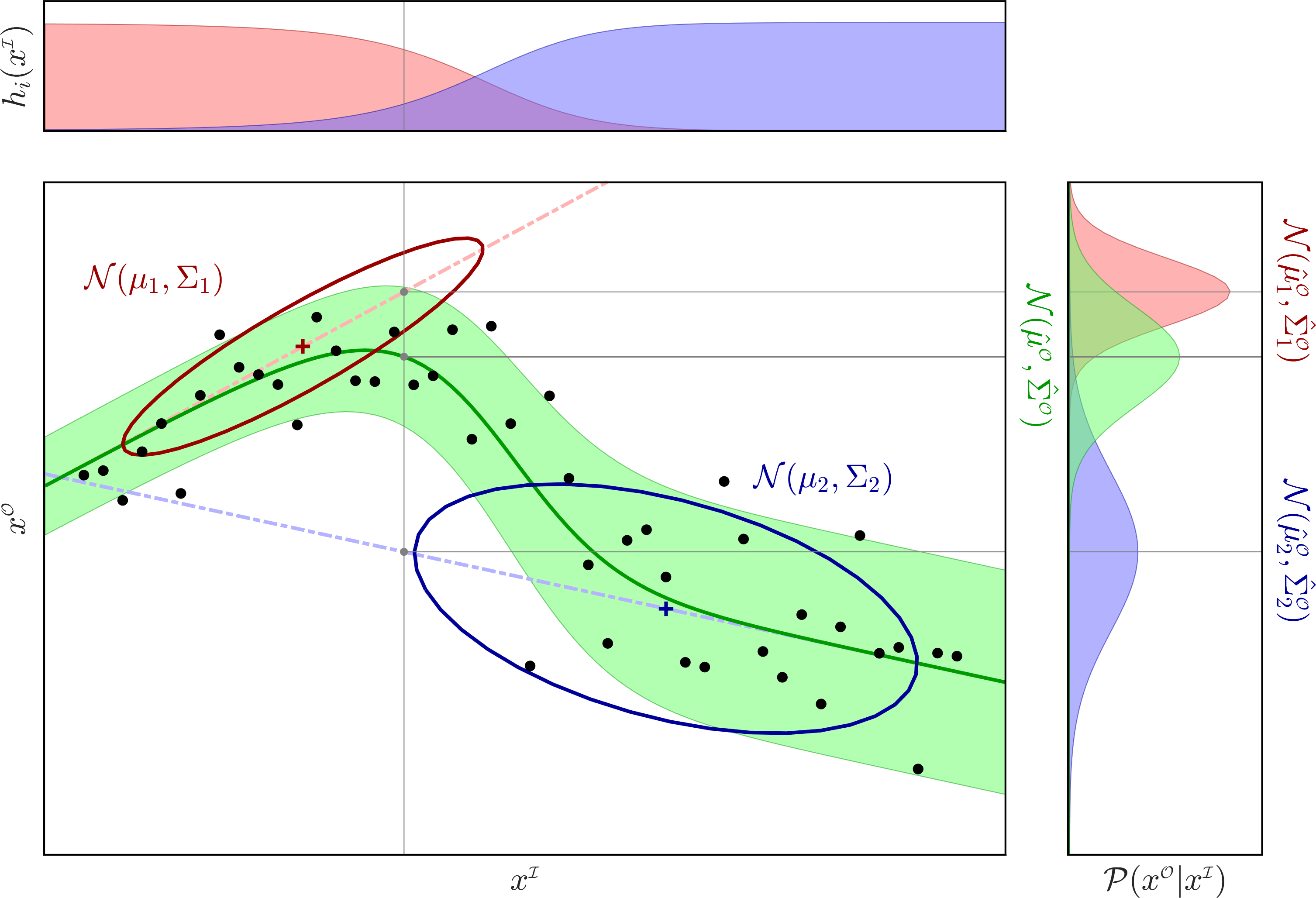}\hspace{3mm}
  \includegraphics[width=.3\columnwidth]{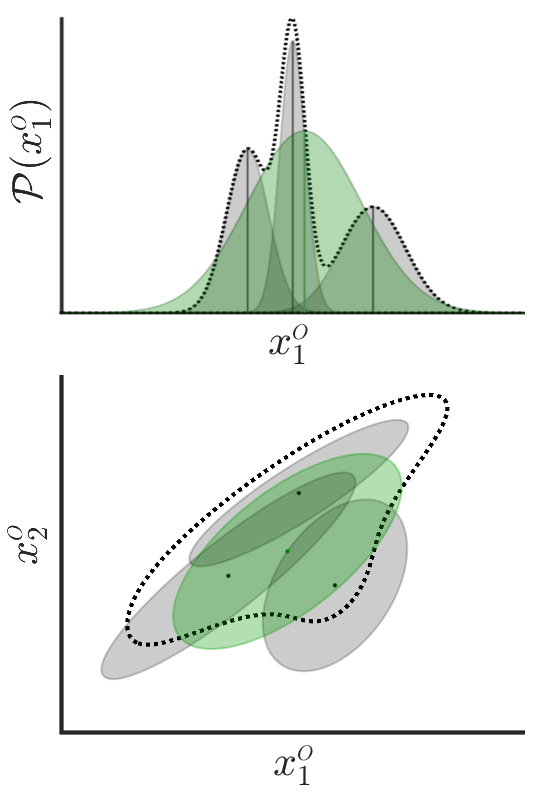}
  \caption{
\emph{Left:} Gaussian mixture regression (GMR) for 1D input $x^\ty{I}$ and 1D output $x^\ty{O}$, with a GMM composed of two Gaussians. 
\emph{Right:} Gaussian that best approximates a mixture of three Gaussians. The multimodal distributions in dashed line depict the probability density functions for the mixtures of three Gaussians in gray color (examples in 1D and 2D are depicted). The Gaussians in green color approximate these multimodal distributions.
  }
  \label{fig:GMR}
\end{figure}

\noindent Gaussian mixture regression (GMR) is a another popular technique for time series and motion representations~\cite{Ghahramani94,Calinon19chapter}. It relies on linear transformation and conditioning properties of multivariate Gaussian distributions. GMR provides a synthesis mechanism to compute output distributions with a computation time independent of the number of datapoints used to train the model. A characteristic of GMR is that it does not model the regression function directly. Instead, it first models the joint probability density of the data in the form of a Gaussian mixture model (GMM). It can then compute the regression function from the learned joint density model, resulting in very fast computation of a conditional distribution. 

In GMR, both input and output variables can be multidimensional. Any subset of input--output dimensions can be selected, which can change, if required, at each time step. Thus, any combination of input--output mappings can be considered, where expectations on the remaining dimensions are computed as a multivariate distribution. 
In the following, we will denote the block decomposition of a datapoint $\bm{x}_t\in\mathbb{R}^D$ at time step $t$, and the center $\bm{\mu}_k$ and covariance $\bm{\Sigma}_k$ of the $k$-th Gaussian in the GMM as
\begin{equation}
	\bm{x}_t=\begin{bmatrix}\bm{x}^\ty{I}_t \\ \bm{x}^\ty{O}_t\end{bmatrix}
	\;,\quad
	\bm{\mu}_k=\begin{bmatrix}\bm{\mu}^\ty{I}_k \\ \bm{\mu}^\ty{O}_k\end{bmatrix}
	\;,\quad
	\bm{\Sigma}_k=\begin{bmatrix}\bm{\Sigma}^\ty{I}_k \;\bm{\Sigma}^\ty{IO}_k \\ \bm{\Sigma}^\ty{OI}_k
	\bm{\Sigma}^\ty{O}_k\end{bmatrix}.
\end{equation}
We first consider the example of time-based trajectories by using $\bm{x}^\ty{I}_t$ as a time variables. At each time step $t$, $\mathcal{P}(\bm{x}^\ty{O}_t|\bm{x}^\ty{I}_t)$ can be computed as the multimodal conditional distribution
\begin{align}
	\mathcal{P}(\bm{x}^\ty{O}_t|\bm{x}^\ty{I}_t)
	&\;=\;
	\sum_{k=1}^K h_k(\bm{x}^\ty{I}_t)\;
	\mathcal{N}\!\left(\bm{\hat{\mu}}^\ty{O}_k(\bm{x}^\ty{I}_t),\bm{\hat{\Sigma}}^\ty{O}_k\right), 
	\label{eq:GMRcondProb}\\
	\mathrm{with}\quad \bm{\hat{\mu}}^\ty{O}_k(\bm{x}^\ty{I}_t)
	&\;=\;
	\bm{\mu}^\ty{O}_k + \bm{\Sigma}^\ty{OI}_k{\bm{\Sigma}^\ty{I}_k}^{-1}
	(\bm{x}^\ty{I}_t-\bm{\mu}^\ty{I}_k)
	\;,
	\nonumber\\
	\bm{\hat{\Sigma}}^\ty{O}_k
	&\;=\; \bm{\Sigma}^\ty{O}_k -
	\bm{\Sigma}^\ty{OI}_k{\bm{\Sigma}^\ty{I}_k}^{-1} \bm{\Sigma}^\ty{IO}_k,
	\nonumber\\
	\mathrm{and}\quad
	h_k(\bm{x}^\ty{I}_t)
	&\;=\;
	\frac{\pi_k \; \mathcal{N}(\bm{x}^\ty{I}_t |\; \bm{\mu}^\ty{I}_k,\bm{\Sigma}^\ty{I}_k)}
	{\sum_{i=1}^K \pi_i \; \mathcal{N}(\bm{x}^\ty{I}_t |\; \bm{\mu}^\ty{I}_i,\bm{\Sigma}^\ty{I}_i)},
	\nonumber
\end{align}
computed with 
\begin{equation*}
	\mathcal{N}(\bm{x}^\ty{I}_t |\; \bm{\mu}^\ty{I}_k,\bm{\Sigma}^\ty{I}_k)
	= (2\pi)^{-\frac{D}{2}} {|\bm{\Sigma}^\ty{I}_k|}^{-\frac{1}{2}} 
	\exp\Big(\!-\frac{1}{2}{(\bm{x}^\ty{I}_t-\bm{\mu}^\ty{I}_k)}^{\trsp} {\bm{\Sigma}^\ty{I}_k}^{-1} 
	(\bm{x}^\ty{I}_t-\bm{\mu}^\ty{I}_k)\!\Big).
\end{equation*}

When a unimodal output distribution is required, the law of total mean and variance (see Fig.~\ref{fig:GMR}-\emph{right}) can be used to approximate the distribution with the Gaussian 
\begin{align}
	\mathcal{P}(\bm{x}_t^\ty{O} | \bm{x}_t^\ty{I})
	&= \mathcal N \Big(\bm{x}_t^\ty{O} |\; \bm{\hat\mu}{}^\ty{O}\!(\bm{x}_t^\ty{I}), \bm{\hat\Sigma}{}^\ty{O}\!(\bm{x}_t^\ty{I}) \Big),\\
	\mathrm{with}\quad 
	\bm{\hat{\mu}}{}^\ty{O}\!(\bm{x}_t^\ty{I})
	&= \sum_{k=1}^K \! h_k(\bm{x}_t^\ty{I}) \; \bm{\hat{\mu}}_k^\ty{O}\!(\bm{x}_t^\ty{I}),\nonumber\\
	\mathrm{and} \quad
	\bm{\hat\Sigma}{}^\ty{O}\!(\bm{x}_t^\ty{I})
	&= \sum_{k=1}^K \! h_k(\bm{x}^\ty{I}_t) \Big( \bm{\hat{\Sigma}}_k^\ty{O} \!+\!
	\bm{\hat{\mu}}_k^\ty{O}\!(\bm{x}_t^\ty{I}) \; {\bm{\hat{\mu}}_k^\ty{O}\!(\bm{x}_t^\ty{I})}^\trsp \Big) -
	\bm{\hat{\mu}}{}^\ty{O}\!(\bm{x}_t^\ty{I}) \, {\bm{\hat{\mu}}{}^\ty{O}\!(\bm{x}_t^\ty{I})\phantom{}}^\trsp. \nonumber
\end{align}
Figure \ref{fig:GMR} presents an example of GMR with 1D input and 1D output. With the GMR representation, LWR corresponds to a GMM with diagonal covariances. Expressing LWR in the more general form of GMR has several advantages: (1) it allows the encoding of local correlations between the motion variables by extending the diagonal covariances to full covariances; (2) it provides a principled approach to estimate the parameters of the RBFs, similar to a GMM parameters fitting problem; (3) it often allows a significant reduction of the number of RBFs, because the position and spread of each RBF are also estimated; and (4) the (online) estimation of the mixture model parameters and the model selection problem (automatically estimating the number of basis functions) can readily exploit techniques compatible with GMM (Bayesian nonparametrics with Dirichlet processes, spectral clustering, small variance asymptotics, expectation-maximization procedures, etc.). 

Another approach to encode and synthesize a movement is to rely on time-invariant autonomous systems. GMR can also be employed in this context to retrieve an autonomous system $\mathcal{P}(\bm{\dot{x}}|\bm{x})$ from the joint distribution $\mathcal{P}(\bm{x},\bm{\dot{x}})$ encoded in a GMM, where $\bm{x}$ and $\bm{\dot{x}}$ are position and velocity, respectively (see~\cite{Hersch08TRO} for details). Similarly, it can be used in an autoregressive context by retrieving $\mathcal{P}(\bm{x}_t|\bm{x}_{t-1},\bm{x}_{t-2},\ldots,\bm{x}_{t-T})$ at each time step $t$, from the joint encoding of the positions on a time window of size $T$.

Practical applications of GMR include the analysis of speech signals~\cite{Toda07,Hueber16}, electromyography signals~\cite{Jaquier17IROS}, vision and MoCap data~\cite{Tian13}, and cancer prognosis~\cite{Falk06}. A Matlab code example \texttt{demo\_GMR01.m} can be found in~\cite{pbdlib}.

\subsection{Bernstein basis functions}
\label{sec:Bernstein}

\begin{figure}
  \centering
  \includegraphics[width=.9\columnwidth]{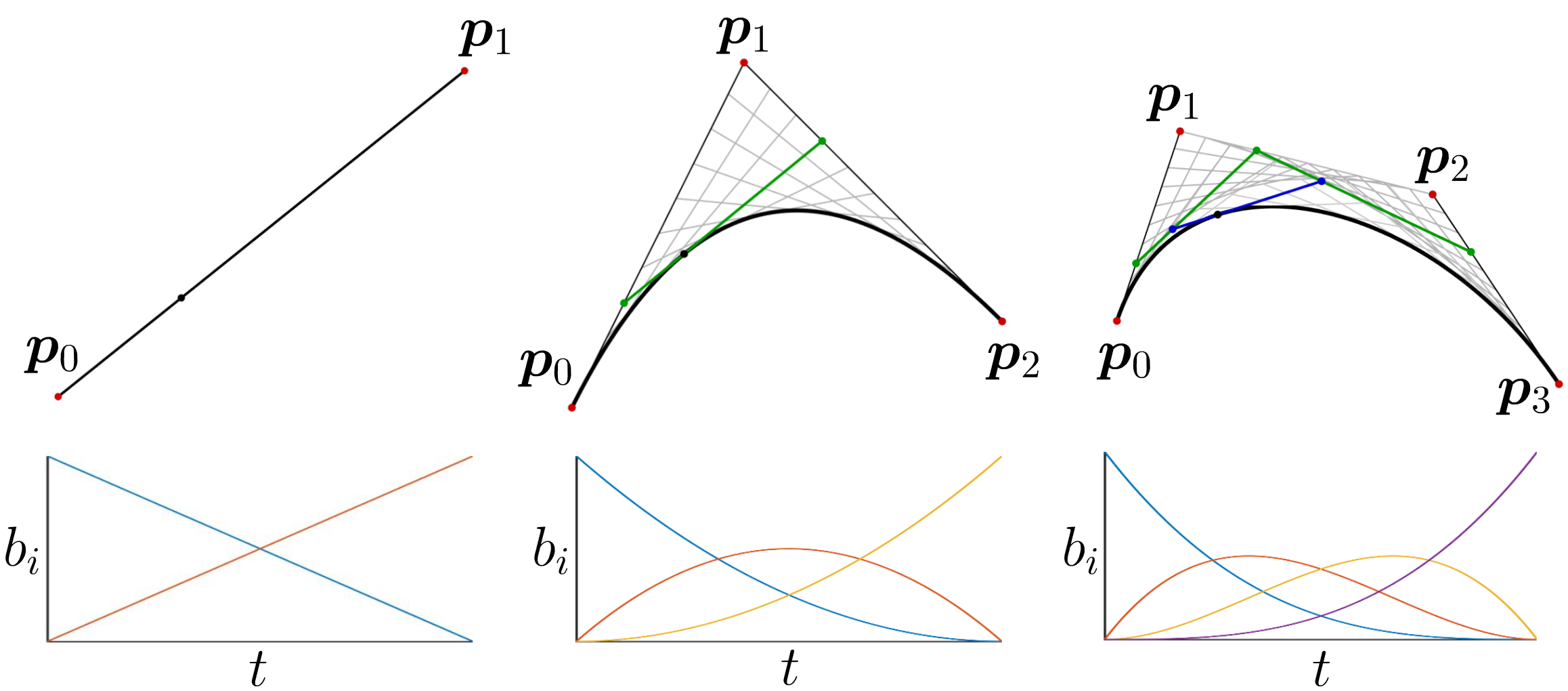}
  \caption{
Linear (\emph{left}), quadratic (\emph{center}) and cubic (\emph{right}) B\'ezier curves constructed as a weighted superposition of Bernstein basis functions.
  }
  \label{fig:Bezier}
\end{figure}

\noindent B\'ezier curves are well-known representations of trajectories~\cite{Farouki12}. Their underlying representation is a superposition of basis functions, which is overlooked in many applications. For $0\leqslant t\leqslant 1$, a linear B\'ezier curve is the line traced by the function $\bm{x}_\ty{\bm{p}_0,\bm{p}_1}(t)$, from $\bm{p}_0$ to $\bm{p}_1$,
\begin{align}
	\bm{x}_\ty{\bm{p}_0,\bm{p}_1}(t) &= (1-t) \, \bm{p}_0 + t \, \bm{p}_1.
\end{align}
For $0\leqslant t\leqslant 1$, a quadratic B\'ezier curve is the path traced by the function 
\begin{align}
	\bm{x}_\ty{\bm{p}_0,\bm{p}_1,\bm{p}_2}(t) &= (1-t) \; \bm{x}_\ty{\bm{p}_0,\bm{p}_1}(t) + t \; \bm{x}_\ty{\bm{p}_1,\bm{p}_2}(t)\nonumber\\
	&= (1-t) \Big((1-t) \bm{p}_0 + t \, \bm{p}_1\Big) + t \Big((1-t) \bm{p}_1 + t \, \bm{p}_2\Big)\nonumber\\
	&= (1-t)^{2} \bm{p}_0 + 2(1-t)t \, \bm{p}_1 + t^{2} \bm{p}_2.
\end{align}
For $0\leqslant t\leqslant 1$, a cubic B\'ezier curve is the path traced by the function 
\begin{align}
	\bm{x}_\ty{\bm{p}_0,\bm{p}_1,\bm{p}_2,\bm{p}_3}(t) &= (1-t) \; \bm{x}_\ty{\bm{p}_0,\bm{p}_1,\bm{p}_2}(t) + t \;  \bm{x}_\ty{\bm{p}_1,\bm{p}_2,\bm{p}_3}(t)\nonumber\\
	&= (1-t)^{3} \bm{p}_0 + 3(1-t)^2 t \bm{p}_1 + 3(1-t)t^2 \bm{p}_2 + t^3 \bm{p}_3.
\end{align}
For $0\leqslant t\leqslant 1$, a recursive definition for a B\'ezier curve of degree $n$ can be expressed as a linear interpolation of a pair of corresponding points in two B\'ezier curves of degree $n-1$, namely
\begin{equation}
	\bm{x}(t) = \sum_{i=0}^n b_{i,n}(t) \, \bm{p}_i,
	\quad\mathrm{with}\quad
	b_{i,n}(t) = \frac{n!}{i!(n-i)!} \; (1-t)^{n-i} \; t^i,
\end{equation}
with $b_{i,n}(t)$ the Bernstein basis polynomials of degree n, where $\frac{n!}{i!(n-i)!}$ are binomial coefficients, which can also be noted as $\binom{n}{i}$. 

Figure \ref{fig:Bezier} illustrates the construction of B\'ezier curves of different orders. 
Practical applications are diverse but include most notably trajectories in computer graphics~\cite{Farouki12} and path planning~\cite{Egerstedt10}.
A Matlab code example \texttt{demo\_Bezier01.m} can be found in~\cite{pbdlib}.

\subsection{Fourier basis functions}
\label{sec:Fourier}
In this section, we will adopt a notation to make links with the superposition of basis functions seen in Fig.~\ref{fig:superposition}. By starting with the unidimensional case, we will consider a signal $g(x)$ varying along a variable $x$, where $x$ will be used as a generic variable that can for example be a time variable as in the example of Fig.~\ref{fig:superposition}, or the coordinates of a pixel in an image. The signal $g(x)$ can be approximated as a weighted superposition of basis functions with 
\begin{align*}
	g(x) &= \sum_{k=-K\!+\!1}^{K\!-\!1} w_k \, \phi_k(x)\\
	&= \bm{w}^\trsp \bm{\phi}(x), 
\end{align*}
where $\bm{w}$ and $\bm{\phi}(x)$ are vectors formed with the elements of $w_k$ and $\phi_k(x)$, respectively.
$w_k$ and $\phi_k(x)$ denote the coefficients and basis functions of the Fourier series, with
\begin{align}
	\phi_k(x) &= \frac{1}{L} \exp\!\left(-i\frac{2\pi k x}{L}\right)\nonumber\\
	&= \frac{1}{L} \Bigg(\!\cos\!\left(\frac{2\pi k x}{L}\right) - i \, \sin\!\left(\frac{2\pi k x}{L}\right) \!\Bigg)
	, \;\forall k\!\in\![-K\!+\!1,\ldots,K\!-\!1],
	\label{eq:complExp1D}
\end{align}
with $i$ the imaginary unit of a complex number ($i^2=-1$).

In time series encoding, the use of Fourier basis functions provides useful connections between the spatial domain and the frequency domain. In the context of Gaussian mixture models, several Fourier series properties can be exploited, notably regarding zero-centered Gaussians, shift, symmetry, and linear combination. 
These properties are reported in Table \ref{tab:Fourier} for the 1D case.

\begin{table*}
\caption{Fourier series properties (1D case).}
\label{tab:Fourier}
\begin{mdframed}

\noindent\textbf{Symmetry property:}\\
If $g(x)$ is real and even, $\phi_k(x)$ in \eqref{eq:complExp1D} is also real and even, simplifying to $\phi_k(x) = \frac{1}{L} \cos\!\left(\frac{2\pi k x}{L}\right)$, which then, in practice, only needs an evaluation on the range $k\!\in\![0,\ldots,K\!-\!1]$, as the basis functions are even. We then have $g(x) = w_0 + \sum_{k=1}^{K\!-\!1} w_k \, 2\cos\!\left(\frac{2\pi k x}{L}\right)$, by exploiting $\cos(0)\!=\!1$.\\[-2mm]

\noindent\textbf{Shift property:}\\
If $w_k$ are the Fourier series coefficients of a function $g(x)$, $\exp(-i \frac{2\pi k \mu}{L}) w_k$ are the Fourier coefficients of $g(x-\mu)$.\\[-2mm]

\noindent\textbf{Combination property:}\\
If $w_{k,1}$ (resp.\ $w_{k,2}$) are the Fourier series coefficients of a function $g_1(x)$ (resp.\ $g_2(x)$), then $\alpha_1 w_{k,1}+\alpha_2 w_{k,2}$ are the Fourier coefficients of $\alpha_1 g_1(x)+\alpha_2 g_2(x)$.\\[-2mm]

\noindent\textbf{Gaussian property:}\\
If $g_0(x)=\mathcal{N}(x \,|\, 0, \sigma^2)=(2\pi\sigma^2)^{-\frac{1}{2}} \exp(-\frac{x^2}{2\sigma^2})$
is mirrored to create a real and even periodic function $g(x)$ of period $L\gg\sigma$ (implementation details will follow), the corresponding Fourier series coefficients are of the form $w_k=\exp(-\frac{2\pi^2 k^2 \sigma^2}{L^2})$.\\[-2mm]

\end{mdframed}
\end{table*}

Well-known applications of Fourier basis functions in the context of time series include speech processing~\cite{Toda07,Hueber16} and the analysis of periodic motions such as gaits~\cite{Antonsson85}. Such decompositions also have a wider scope of applications, as illustrated next with ergodic control. 

\subsection{Ergodic control} 

\begin{figure}
  \centering
  \includegraphics[width=\columnwidth]{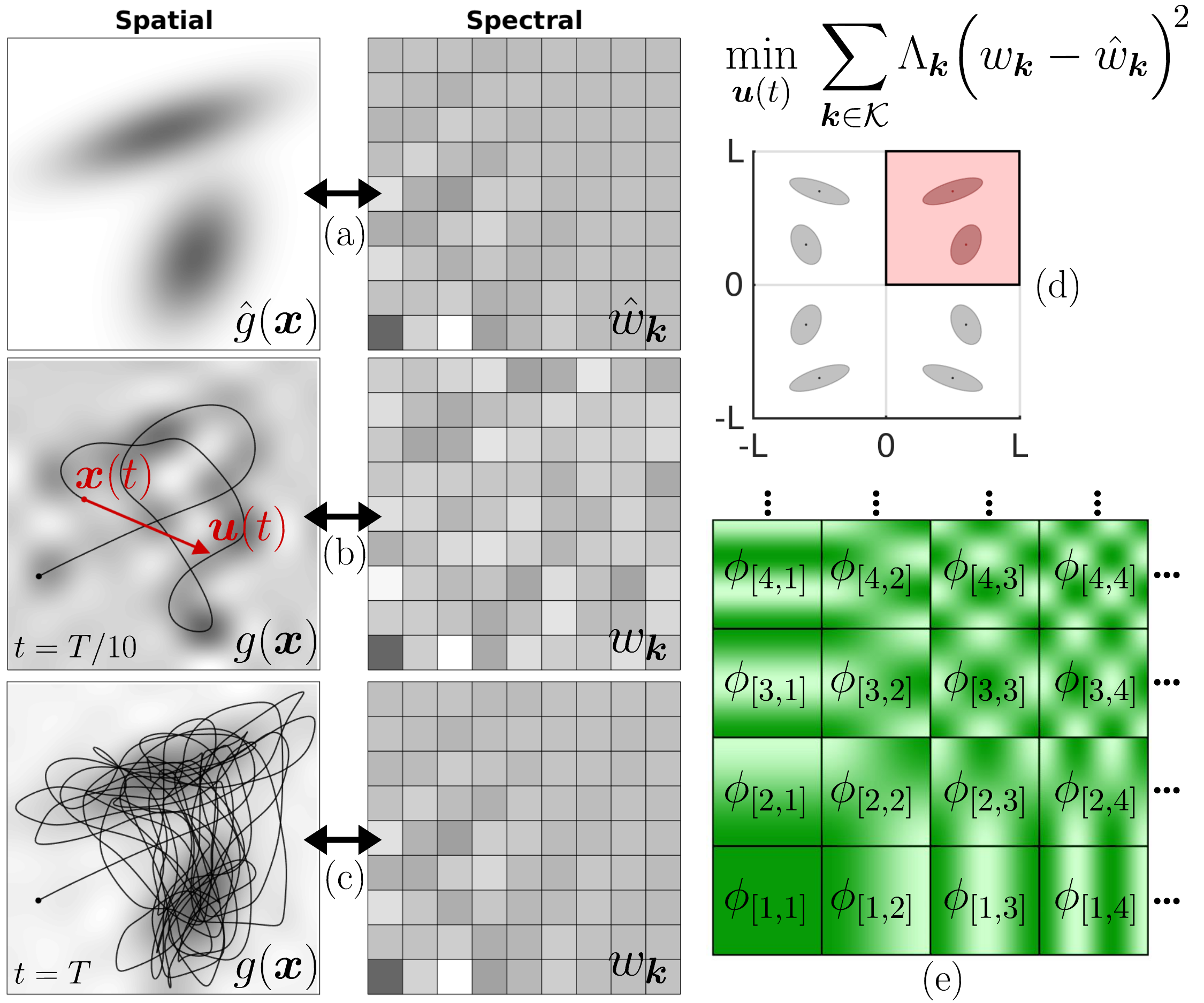}
  \caption{
2D ergodic control problem. 
\textbf{(a)} shows the spatial distribution $\hat{g}(\bm{x})$ that the agent has to explore, encoded here as a mixture of two Gaussians (gray colormap in left graph). The right graphs show the corresponding Fourier series coefficients $\hat{w}_{\bm{k}}$ in the frequency domain ($K=9$ coefficients per dimension), which can be computed analytically by exploiting the shift, symmetry and linear combination properties of Gaussians. 
\textbf{(b)} shows the evolution of the reconstructed spatial distribution $g(\bm{x})$ (left graph) and the computation of the next control command $\bm{u}$ (red arrow) after $T/10$ iterations. The corresponding Fourier series coefficients $w_{\bm{k}}$ are shown in the right graph. 
\textbf{(c)} shows that after $T$ iterations, the agent covers the space in proportion to the desired spatial distribution, with a good match of coefficients in the frequency domain (we can see that $\hat{w}_{\bm{k}}$ and $w_{\bm{k}}$ are nearly the same). 
\textbf{(d)} shows how a periodic signal $\hat{g}(\bm{x})$ (with range $[-L/2,L/2]$ for each dimension) can be constructed from the original mixture of two Gaussians $\hat{g}_0(\bm{x})$ (red area). The constructed signal $\hat{g}(\bm{x})$ is composed of eight Gaussians in this 2D example (mirroring the Gaussians along horizontal and vertical axes to construct an even signal of period $L$). 
\textbf{(e)} depicts the first few basis functions of the Fourier series (for the first four coefficients in each dimension), represented as a 2D colormap corresponding to periodic signals of different frequencies along two axes. 
  }
  \label{fig:ergodic}
\end{figure}

\noindent In ergodic control, the aim is to find a series of control commands $\bm{u}(t)$ so that the retrieved trajectory $\bm{x}(t)\in\mathbb{R}^D$ covers a bounded space $\mathcal{X}$ in proportion of a desired spatial distribution $\hat{g}(\bm{x})$, see Fig.~\ref{fig:ergodic}-\emph{(a)}. As proposed in~\cite{Mathew11}, this can be achieved by defining a metric in the spectral domain, by decomposing in Fourier series coefficients both the desired spatial distribution $\hat{g}(\bm{x})$ and the (partially) retrieved trajectory $\bm{x}(t)$.
The goal of ergodic control is to minimize
\begin{align}
	\epsilon &= \frac{1}{2} \sum_{{\bm{k}}\in\mathcal{K}} \Lambda_{\bm{k}} \Big( w_{\bm{k}} - \hat{w}_{\bm{k}} \Big)^{\!2} \label{eq:ergodicerror}\\
	&= \frac{1}{2} {\Big( \bm{w} - \bm{\hat{w}} \Big)}^{\!\trsp} \bm{\Lambda} \, \Big( \bm{w} - \bm{\hat{w}} \Big),
\end{align}
where $\Lambda_{\bm{k}}$ are weights, $\hat{w}_{\bm{k}}$ are the Fourier series coefficients of $\hat{g}(\bm{x})$, and $w_{\bm{k}}$ are the Fourier series coefficients along the trajectory $\bm{x}(t)$. $\mathcal{K}$ is a set of index vectors in $\mathbb{N}^D$ covering the $D$-dimensional array $\bm{k}=\bm{r}\times\bm{r}\times\cdots\times\bm{r}$, with $\bm{r}=[0,1,\ldots,K\!-\!1]$ and $K$ the resolution of the array.\footnote{For $D=2$ and $K=2$, we have $\mathcal{K}=\big\{\left[\begin{smallmatrix}0\\0\end{smallmatrix}\right],\left[\begin{smallmatrix}0\\1\end{smallmatrix}\right],\left[\begin{smallmatrix}1\\0\end{smallmatrix}\right],\left[\begin{smallmatrix}1\\1\end{smallmatrix}\right]\big\}$.}
$\bm{w}\in\mathbb{R}^{K^D}$ and $\bm{\hat{w}}\in\mathbb{R}^{K^D}$ are vectors composed of elements $w_{\bm{k}}$ and $\hat{w}_{\bm{k}}$, respectively. $\bm{\Lambda}\in\mathbb{R}^{K^D\times K^D}$ is a diagonal weighting matrix with elements $\Lambda_{\bm{k}}$.
In \eqref{eq:ergodicerror}, the weights
\begin{equation}
	\Lambda_{\bm{k}} = \left(1+ \|\bm{k}\|^2\right)^{\!-\frac{D+1}{2}}
\end{equation}
assign more importance on matching low frequency components (related to a metric for Sobolev spaces of negative order). 
The Fourier series coefficients $w_{\bm{k}}$ along a trajectory $\bm{x}(t)$ of continuous duration $t$ are defined as
\begin{equation}
	w_{\bm{k}} = \frac{1}{t} \int_{\tau=0}^t \phi_{\bm{k}}\big(\bm{x}(\tau)\big) \; d\tau,
	\label{eq:ck}
\end{equation}
whose discretized version can be computed recursively at each discrete time step $t$ to build
\begin{equation}
	w_{\bm{k}} = \frac{1}{t} \sum_{s=1}^t \phi_{\bm{k}}(\bm{x}_s), 
\end{equation}
or equivalently in vector form $\bm{w} = \frac{1}{t}\sum_{s=1}^t\bm{\phi}(\bm{x}_s)$. 

For a spatial signal $\bm{x}\in\mathbb{R}^D$, where $x_d$ is on the interval $[-\frac{L}{2},\frac{L}{2}]$ of period $L$, $\forall d\!\in\!\{1,\ldots,D\}$, the basis functions of the Fourier series with complex exponential functions are defined as (see Fig.~\ref{fig:ergodic}-\emph{(e)})
\begin{align}
	\phi_{\bm{k}}(\bm{x}) &= \frac{1}{L^D} \prod_{d=1}^D \exp\!\left(-i\frac{2\pi k_d x_d}{L}\right)\nonumber\\
	&= \frac{1}{L^D} \prod_{d=1}^D \cos\!\left(\frac{2\pi k_d x_d}{L}\right) - i \, \sin\!\left(\frac{2\pi k_d x_d}{L}\right)
	, \quad\forall\bm{k}\!\in\!\mathcal{K}.
	\label{eq:complExpND}
\end{align}

\subsubsection*{Computation of Fourier series coefficients $\hat{w}_{\bm{k}}$ for a spatial distribution represented as a Gaussian mixture model}
We consider a desired spatial distribution $\hat{g}_0(\bm{x})$ represented as a mixture of $J$ Gaussians with centers $\bm{\mu}_j$, covariance matrices $\bm{\Sigma}_j$, and mixing coefficients $\alpha_j$ (with $\sum_{j=1}^J\alpha_j=1$ and $\alpha_j\geqslant0$),
\begin{align}
	\hat{g}_0(\bm{x}) &= \sum_{j=1}^J \alpha_j \, \mathcal{N}\big(\bm{x} \,|\, \bm{\mu}_j,\bm{\Sigma}_j\big) \label{eq:phiND}\\
	&= \sum_{j=1}^J \alpha_j \, (2\pi)^{-\!\frac{D}{2}} \, {|\bm{\Sigma}_j|}^{-\!\frac{1}{2}} \,
	\exp\!\Big(\!-\!\frac{1}{2}{(\bm{x}\!-\!\bm{\mu}_j)}^{\trsp} \bm{\Sigma}_j^{-1} (\bm{x}\!-\!\bm{\mu}_j)\!\Big),\nonumber
\end{align}
with each dimension on the interval $[0,\frac{L}{2}]$.
$\hat{g}_0(\bm{x})$ is extended to a periodized function by constructing an even function on the interval $\mathcal{X}$, where each dimension $x_d$ is on the interval $\mathcal{X}\!=\![-\frac{L}{2},\frac{L}{2}]$ of period $L$. This is achieved with mirror symmetries of the Gaussians around all zero axes, see Fig.~\ref{fig:ergodic}-\emph{(d)}. The resulting spatial distribution can be expressed as a mixture of $2^DJ$ Gaussians
\begin{equation}
	\hat{g}(\bm{x}) = \sum_{j=1}^{J} \sum_{m=1}^{2^D} \frac{\alpha_j}{2^D} \; \mathcal{N}\big(\bm{x} \,\big|\, \bm{A}_m\bm{\mu}_j,\bm{A}_m\bm{\Sigma}_j\bm{A}_m^\trsp\big),
	\label{eq:phiNDm}
\end{equation}
with linear transformation matrices $\bm{A}_m$.\footnote{$\bm{A}_m\!=\!\diag(\bm{H}_{2^D-D+1:2^D,m})$, where $\bm{H}_{2^D-D+1:2^D,m}$ is a vector composed of the last $D$ elements in the column $m$ of the Hadamard matrix $\bm{H}$ of size $2^D$. Alternatively, $\bm{A}_m\!=\!\diag\big(\text{vec}(\bm{\ell}_m)\big)$ can be constructed with the array $\bm{\ell}_m$, with $m$ indexing the first dimension of the array $\bm{\ell}\!=\!\bm{s}\times\bm{s}\times\cdots\times\bm{s} \in\mathbb{Z}^{2\!\times\!2\!\times\!\ldots\!\times\!2}$ with $\bm{s}\!=\![-1,1]$. In 2D, we have $\bm{A}_1\!=\!\left[\begin{smallmatrix}-1&0\\0&-1\end{smallmatrix}\right]$, $\bm{A}_2\!=\!\left[\begin{smallmatrix}-1&0\\0&1\end{smallmatrix}\right]$, $\bm{A}_3\!=\!\left[\begin{smallmatrix}1&0\\0&-1\end{smallmatrix}\right]$ and $\bm{A}_4\!=\!\left[\begin{smallmatrix}1&0\\0&1\end{smallmatrix}\right]$, see Fig.~\ref{fig:ergodic}-\emph{(d)}.} 
By exploiting the symmetry, shift and Gaussian properties presented in Section~\ref{sec:Fourier}, the Fourier series coefficients $\hat{w}_{\bm{k}}$ can be analytically computed as
\begin{align}
	\hat{w}_{\bm{k}} &= \int_{\bm{x}\in\mathcal{X}} \hat{g}(\bm{x}) \; \phi_{\bm{k}}(\bm{x}) \; \text{d}\bm{x} \nonumber\\
	&=  \frac{1}{L^D} \sum_{j=1}^J \sum_{m=1}^{2^D} \frac{\alpha_j}{2^D} \exp\!\left(-i\frac{2\pi \bm{k}^{\trsp}\bm{A}_m \bm{\mu}_j}{L}\right) 
	\exp\!\left(-\frac{2\pi^2 \bm{k}^{\trsp} \bm{A}_m \bm{\Sigma}_j \bm{A}_m^\trsp \bm{k}}{L^2}\right) \nonumber\\
	&=  \frac{1}{L^D} \sum_{j=1}^J \sum_{m=1}^{2^{D-1}} \frac{\alpha_j}{2^{D-1}} \cos\!\left(\frac{2\pi \bm{k}^{\trsp} \bm{A}_m \bm{\mu}_j}{L}\right) 
	\exp\!\left(-\frac{2\pi^2 \bm{k}^{\trsp} \bm{A}_m \bm{\Sigma}_j \bm{A}_m^\trsp \bm{k}}{L^2}\right).
	\label{eq:phikND}
\end{align}
With this mirroring, we can see that $\hat{w}_{\bm{k}}$ are real and even, where an evaluation over $\bm{k}\!\in\!\mathcal{K}$, $j\!\in\!\{1,2,\ldots,J\}$ and $m\!\in\!\{1,2,\ldots,2^{D-1}\}$ in \eqref{eq:phikND} is sufficient to fully characterize the signal.

\subsubsection*{Controller for a spatial distribution represented as a Gaussian mixture model}
In~\cite{Mathew11}, ergodic control is set as the constrained problem of computing a control command $\bm{\hat{u}}(t)$ at each time step $t$ with 
\begin{equation}
	\bm{\hat{u}}(t) = \arg\min_{\bm{u}(t)}\; \epsilon\big(\bm{x}(t\!+\!\Delta t)\big), 
	\quad \text{s.t.} \quad
	\bm{\dot{x}}(t) = f\big(\bm{x}(t),\bm{u}(t)\big),\quad \|\bm{u}(t)\| \leqslant u^{\max},
	\label{eq:objFct}
\end{equation}
where the simple system $\bm{\dot{x}}(t)=\bm{u}(t)$ is considered (control with velocity commands), and where the error term is approximated with the Taylor series
\begin{equation}
	\epsilon\big(\bm{x}(t\!+\!\Delta t)\big) \;\approx\; \epsilon\big(\bm{x}(t)\big) \;+\; \dot{\epsilon}\big(\bm{x}(t)\big) \Delta t 
	\;+\; \frac{1}{2} \ddot{\epsilon}\big(\bm{x}(t)\big) {\Delta t}^2.
	\label{eq:etpp}
\end{equation}
By using \eqref{eq:ergodicerror}, \eqref{eq:ck}, \eqref{eq:complExpND} and the chain rule $\frac{\partial f}{\partial t}=\frac{\partial f}{\partial \bm{x}} \frac{\partial \bm{x}}{\partial t}$, the Taylor series
is composed of the control term $\bm{u}(t)$ and $\nabla_{\!\!\bm{x}} \phi_{\bm{k}}\big(\bm{x}(t)\big)\in\mathbb{R}^{1\times D}$, the gradient of $\phi_{\bm{k}}\big(\bm{x}(t)\big)$ with respect to $\bm{x}(t)$.
Solving the constrained objective in \eqref{eq:objFct} then results in the analytical solution (see~\cite{Mathew11} for the complete derivation)
\begin{align}
	\bm{u} = \bm{\tilde{u}}(t)\frac{u^{\max}}{\|\bm{\tilde{u}}(t)\|},
	\quad\mathrm{with}\quad
	\bm{\tilde{u}} &= -\sum_{{\bm{k}}\in\mathcal{K}} \Lambda_{\bm{k}} \big( w_{\bm{k}} - \hat{w}_{\bm{k}} \big) 
	{\nabla_{\!\!\bm{x}} \phi_{\bm{k}}\big(\bm{x}(t)\big)}^\trsp\nonumber\\
	&= -\bm{\nabla}_{\!\!\bm{x}}\bm{\phi}\big(\bm{x}(t)\big) \; \bm{\Lambda} \, \big( \bm{w} - \bm{\hat{w}} \big),
\end{align}
where $\bm{\nabla}_{\!\!\bm{x}}\bm{\phi}\big(\bm{x}(t)\big)\in\mathbb{R}^{D\times K^D}$ is a concatenation of the vectors $\nabla_{\!\!\bm{x}} \phi_{\bm{k}}\big(\bm{x}(t)\big)$.   
Figure \ref{fig:ergodic} shows a 2D example of ergodic control to create a motion approximating the distribution given by a mixture of two Gaussians. A remarkable characteristic of such approach is that the controller produces natural exploration behaviors (see Fig.~\ref{fig:ergodic}-\emph{(c)}) without relying on stochastic noise in the formulation. In the limit case, if the distribution $g(\bm{x})$ is a single Gaussian with a very small isotropic covariance, the controller results in a standard tracking behavior.

Examples of application include surveillance with multi-agent systems~\cite{Mathew11}, active shape estimation~\cite{Abraham17}, and localization for fish-like robots~\cite{Miller16}.
A Matlab code example \texttt{demo\_ergodicControl\_2D01.m} can be found in~\cite{pbdlib}.

\section{Probabilistic movement primitives}
\label{sec:ProMP}

\begin{figure}
  \centering
  \includegraphics[width=.41\columnwidth]{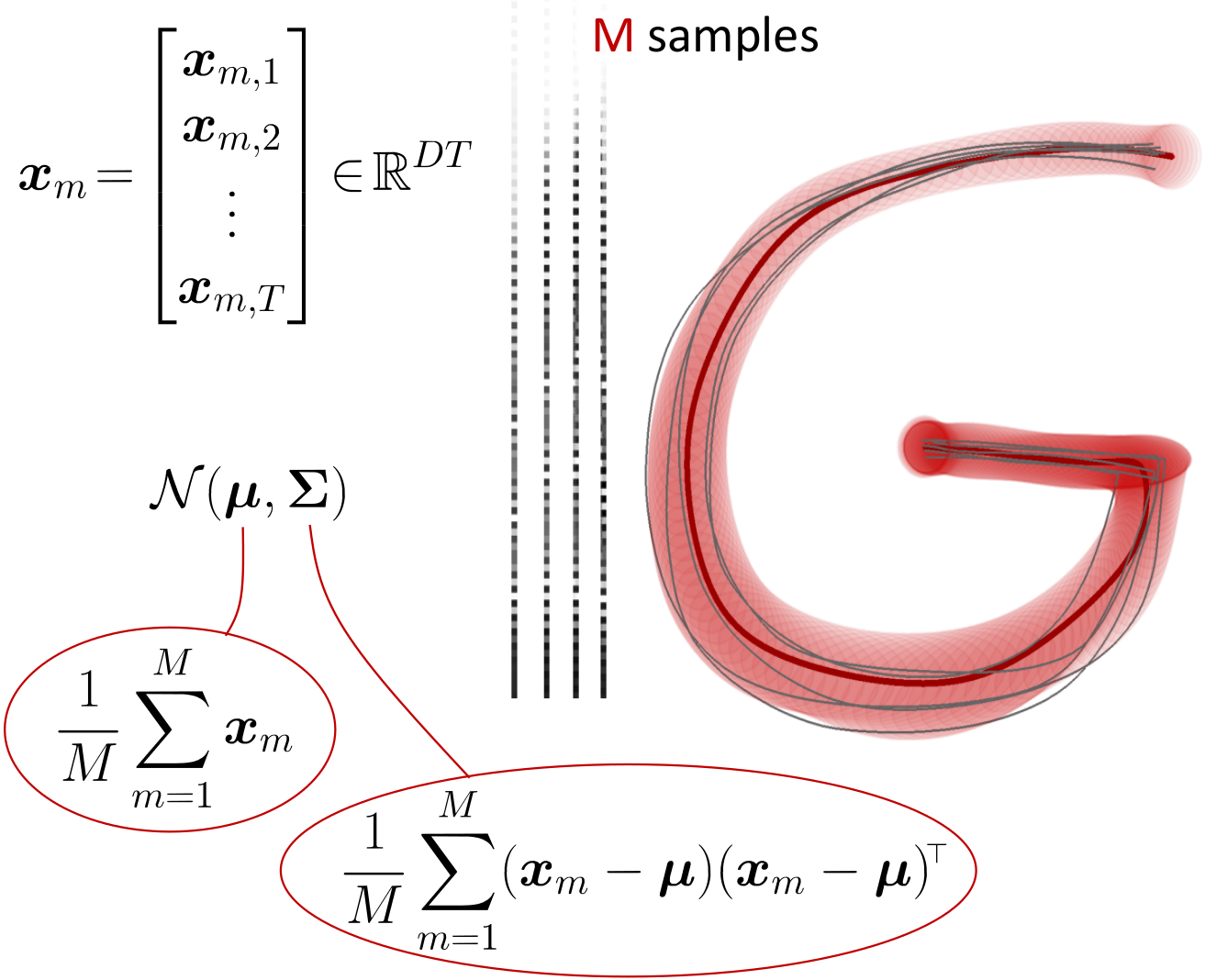}\hspace{4mm}
  \includegraphics[width=.53\columnwidth]{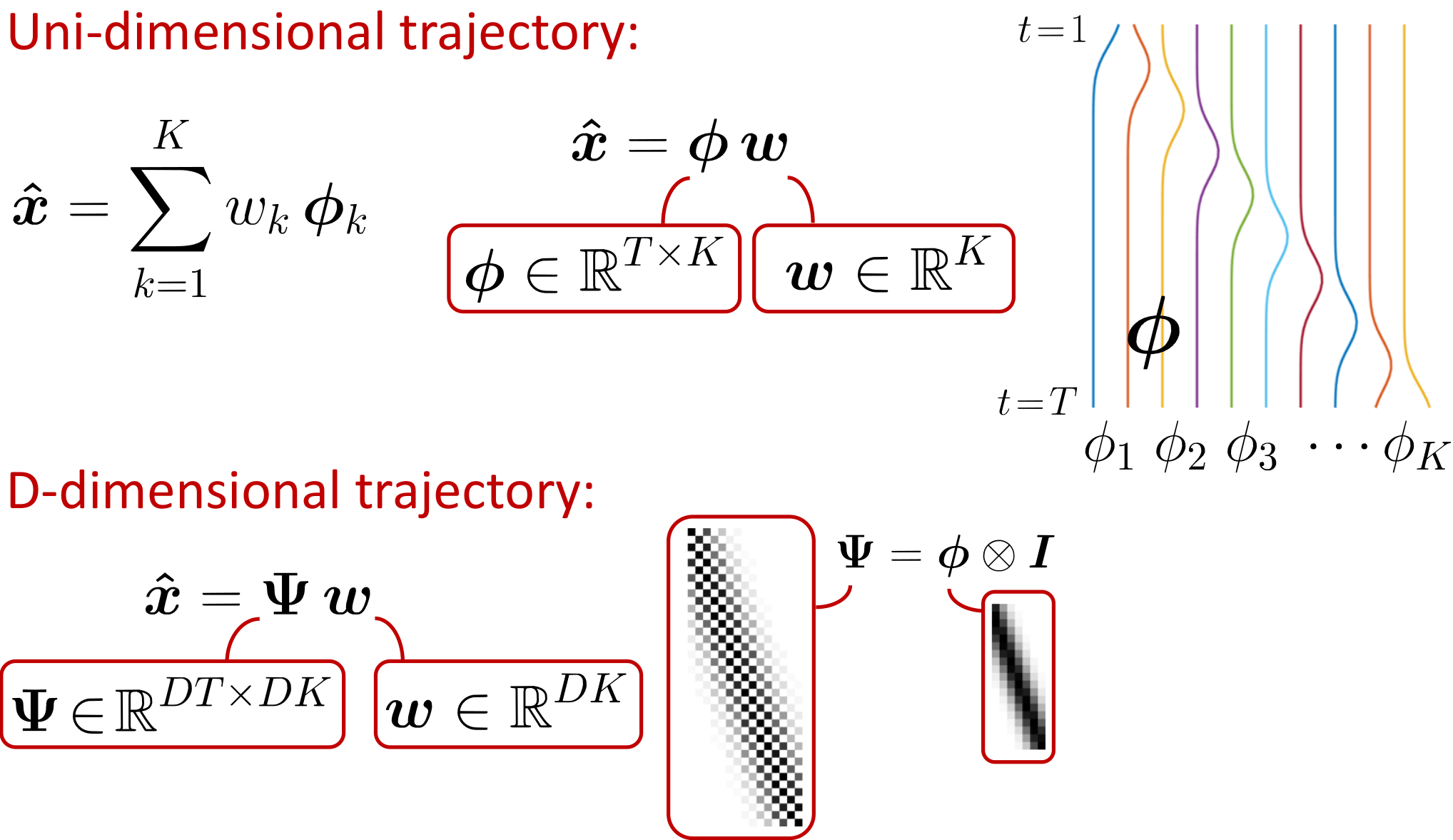}
  \caption{
\emph{Left:} Raw trajectory distribution as a Gaussian of size $DT$ by organizing each of the $M$ samples as a trajectory vector, where each trajectory has $T$ time steps and each point has $D$ dimensions ($T=100$ and $D=2$ in this example). \emph{Right:} Trajectory distribution encoded with probabilistic movement primitives (superposition of $K$ basis functions). The right part of the figure depicts the linear mapping functions $\bm{\phi}$ and $\bm{\Psi}$ created by a decomposition with radial basis functions.
  }
  \label{fig:ProMP1}
\end{figure}

\begin{figure}
  \centering
  \includegraphics[width=.54\columnwidth]{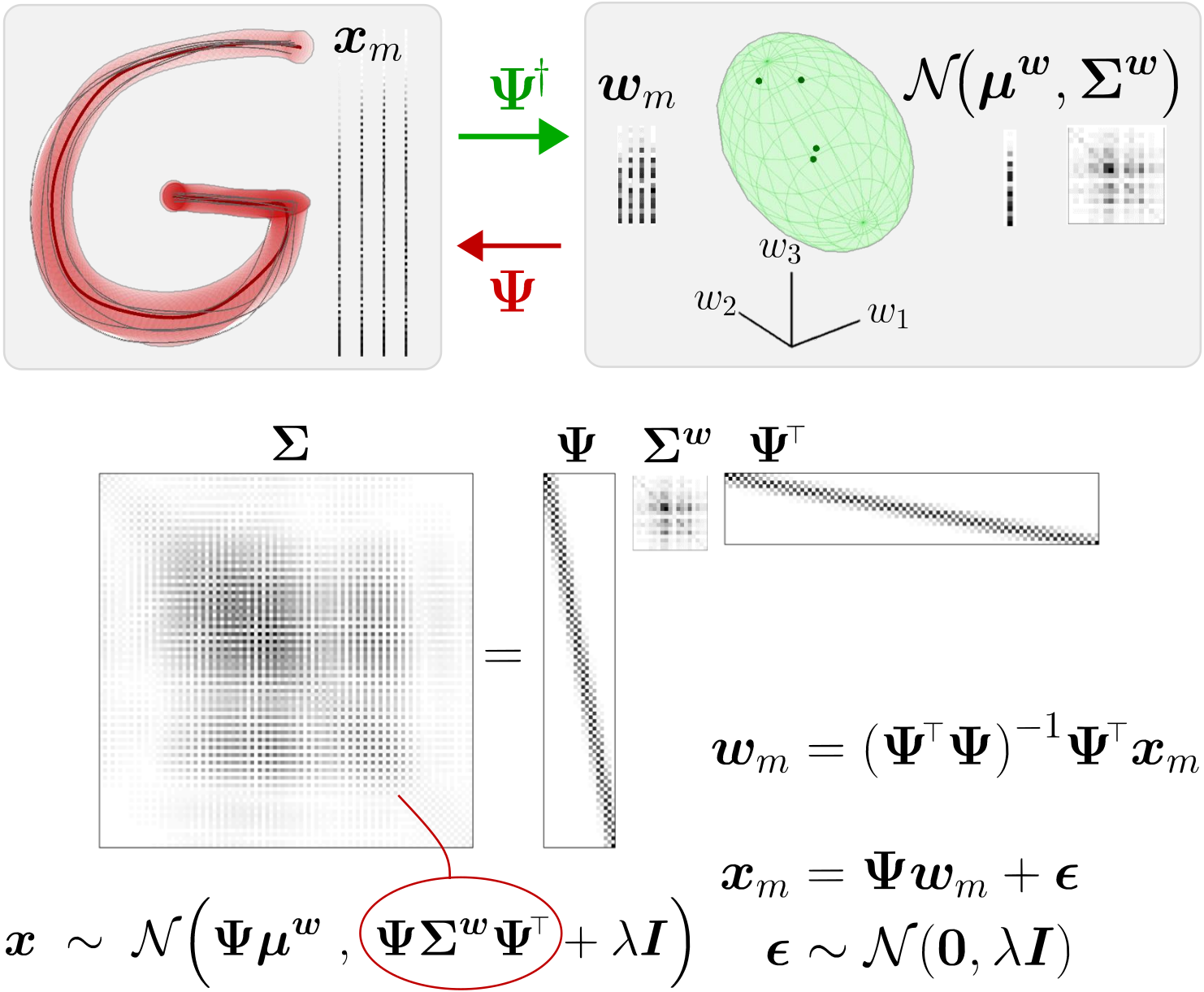}\hspace{4mm}
  \includegraphics[width=.40\columnwidth]{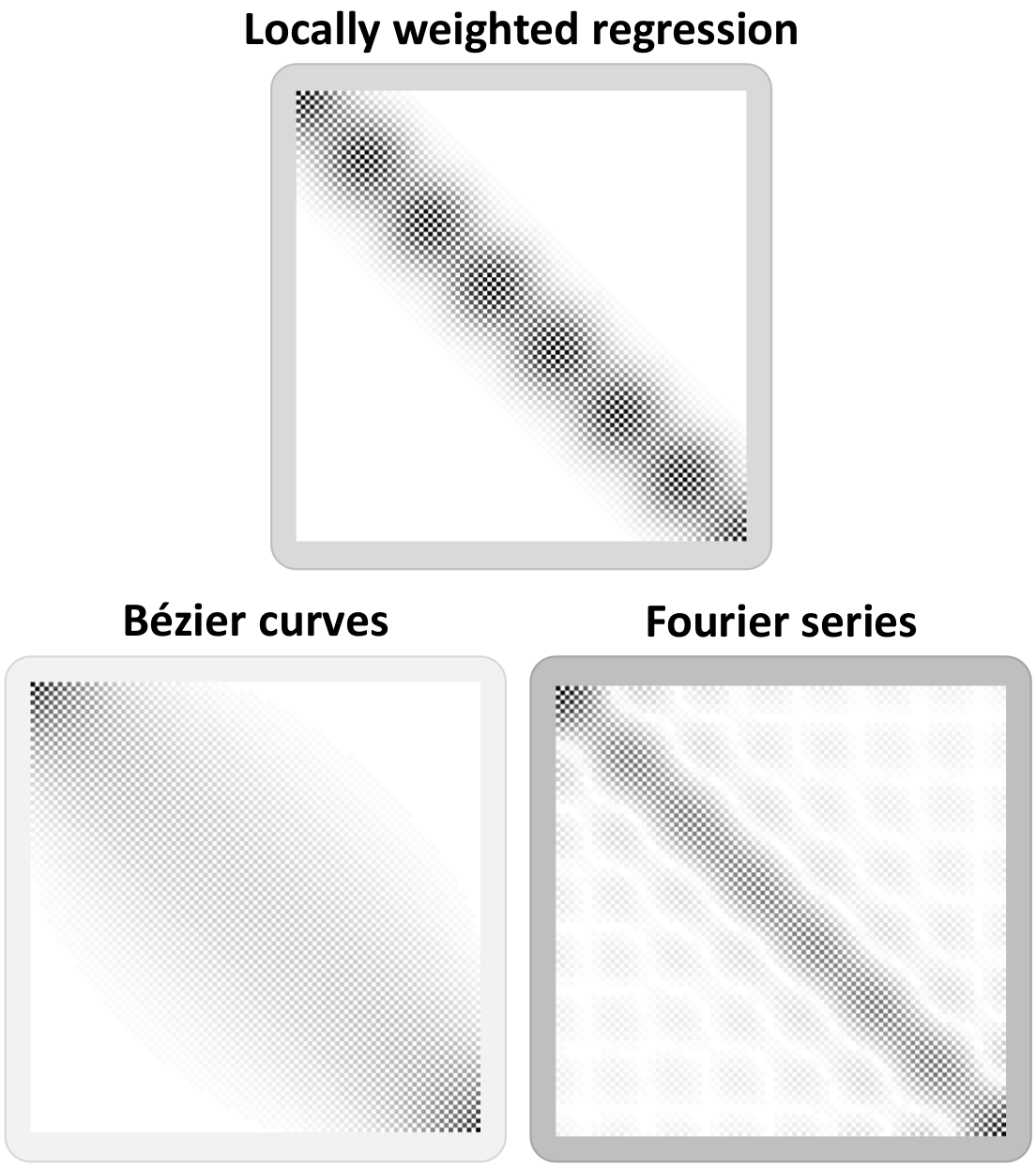}
  \caption{
\emph{Left:} Illustration of probabilistic movement primitives as a linear mapping between the original space of trajectories and a subspace of reduced dimensionality. After projecting each trajectory sample in this subspace (with linear map $\bm{\Psi}^\psin$ computed as the pseudoinverse of $\bm{\Psi}$), a Gaussian is evaluated, which is then projected back to the original trajectory space by exploiting the linear transformation property of multivariate Gaussians (with linear map $\bm{\Psi}$). Such decomposition results in a low rank structure of the covariance matrix, which is depicted in the bottom part of the figure. \emph{Right:} Representation of the covariance matrix $\bm{\Psi} \bm{\Psi}^\trsp$ for various basis functions, all showing some form of sparsity.
  }
  \label{fig:ProMP2}
\end{figure}

\noindent The representation of time series as a superposition of basis functions can also be exploited to construct trajectory distributions. Representing a collection of trajectories in the form of a multivariate distribution has several advantages. First, new trajectories can be stochastically generated. Then, the conditional probability property (see \eqref{eq:GMRcondProb}) can be exploited to generate trajectories passing through via-points (including starting and/or ending points). This is simply achieved by specifying as inputs $\bm{x}^\ty{I}$ in \eqref{eq:GMRcondProb} the datapoints that the system needs to pass through (with corresponding dimensions in the hyperdimensional vector) and by retrieving as output $\bm{x}^\ty{O}$ the remaining parts of the trajectory.

A naive approach to represent a collection of $M$ trajectories in a probabilistic form is to reorganize each trajectory as a hyperdimensional datapoint $\bm{x}_m\!=\!{[\bm{x}_1^\trsp,\bm{x}_2^\trsp,\ldots,\bm{x}_T^\trsp]}^\trsp\!\in\!\mathbb{R}^{DT}$, and fitting a Gaussian $\mathcal{N}(\bm{\mu}^{\bm{x}}, \bm{\Sigma}^{\bm{x}})$ to these datapoints, see Fig.~\ref{fig:ProMP1}-\emph{left}. Since the dimension $DT$ might be much larger than the number of datapoints $M$, a potential solution to this issue could be to consider an eigendecomposition of the covariance (ordered by decreasing eigenvalues) 
\begin{equation}
	\bm{\Sigma}^{\bm{x}} = \bm{V} \bm{D} \bm{V}^\trsp = \sum_{j=1}^{DT} \lambda^2_j \, \bm{v}_j \, \bm{v}_j^\trsp,
\end{equation}
with $\bm{V}\!=\![\bm{v}_1, \bm{v}_2, \ldots, \bm{v}_{DT}]$ and $\bm{D}\!=\!\mathrm{diag}(\lambda^2_1,\lambda^2_2,\ldots,\lambda^2_{DT})$. This can be exploited to project the data in a subspace of reduced dimensionality through principal component analysis. By keeping the first $DK$ components, such approach provides a Gaussian distribution of the trajectories with the structure $\mathcal{N}(\bm{\Psi} \bm{\mu}^{\bm{w}},\bm{\Psi} \bm{\Psi}^\trsp)$, where $\bm{\Psi}\!=\![\bm{v}_1\lambda_1,\, \bm{v}_2\lambda_2,\, \ldots,\, \bm{v}_{DK}\lambda_{DK}]$.

The ProMP (probabilistic movement primitive) model proposed in~\cite{Paraschos13} also encodes the trajectory distribution in a subspace of reduced dimensionality, but provides a RBF structure to this decomposition instead of the eigendecomposition as in the above. It assumes that each sample trajectory $m\!\in\!\{1,\ldots,M\}$ can be approximated by a weighted sum of $K$ normalized RBFs with
\begin{equation}
	\bm{x}_m = \bm{\Psi} \bm{w}_m + \bm{\epsilon},
	\quad\mathrm{where}\quad
	\bm{\epsilon}\sim\mathcal{N}(\bm{0},\sigma^2\bm{I}),
\end{equation}
and basis functions organized as
\begin{equation}
	\bm{\Psi} = \bm{\phi} \otimes \bm{I} = \left[\begin{matrix}
	\bm{I}\phi_1(t_1) & \bm{I}\phi_2(t_1) & \cdots & \bm{I}\phi_K(t_1) \\
	\bm{I}\phi_1(t_2) & \bm{I}\phi_2(t_2) & \cdots & \bm{I}\phi_K(t_2) \\
	\vdots & \vdots & \ddots & \vdots \\
	\bm{I}\phi_1(t_T) & \bm{I}\phi_2(t_T) & \cdots & \bm{I}\phi_K(t_T) 
	\end{matrix}\right],
	\label{eq:Psi}
\end{equation}
with $\bm{\Psi}\!\in\!\mathbb{R}^{DT\times DK}$, identity matrix $\bm{I}\!\in\!\mathbb{R}^{D\!\times\!D}$, and $\otimes$ the Kronecker product operator. A vector $\bm{w}_m\!\in\!\mathbb{R}^{DK}$ can be estimated for each of the $M$ sample trajectories by the least squares estimate
\begin{equation}
	\bm{w}_m = {(\bm{\Psi}^\trsp \bm{\Psi})}^{-1} \bm{\Psi}^\trsp \bm{x}_m.
\end{equation}
By assuming that $\{\bm{w}_m\}_{m=1}^M$ can be represented with a Gaussian $\mathcal{N}(\bm{\mu}^{\bm{w}},\bm{\Sigma}^{\bm{w}})$ characterized by a center $\bm{\mu}^{\bm{w}}\!\in\!\mathbb{R}^{DK}$ and a covariance $\bm{\Sigma}^{\bm{w}}\!\in\!\mathbb{R}^{DK\times DK}$, a trajectory distribution $\mathcal{P}(\bm{x})$ can then be computed as
\begin{equation}
	\bm{x} \;\sim\; \mathcal{N}
	\Big(\bm{\Psi} \bm{\mu}^{\bm{w}}
	\;,\;
	\bm{\Psi} \bm{\Sigma}^{\bm{w}} \bm{\Psi}^\trsp + \sigma^2\bm{I}
	\Big),
	\label{eq:Nw}
\end{equation}
with $\bm{x}\!\in\!\mathbb{R}^{DT}$ a trajectory of $T$ datapoints of $D$ dimensions organized in a vector form and $\bm{I}\!\in\!\mathbb{R}^{DT\!\times\!DT}$, see Figures~\ref{fig:ProMP1} and~\ref{fig:ProMP2}. 

The parameters of the ProMP model are $\sigma^2$, $\mu_k^\ty{I}$, $\Sigma_k^\ty{I}$, $\bm{\mu}^{\bm{w}}$, and $\bm{\Sigma}^{\bm{w}}$. A Gaussian of $DK$ dimensions is estimated, 
providing a compact representation of the movement, separating the temporal components $\bm{\Psi}$ and spatial components $\mathcal{N}(\bm{\mu}^{\bm{w}},\bm{\Sigma}^{\bm{w}})$. Similarly to LWR, ProMP can be coupled with GMM/GMR to automatically estimate the location and bandwidth of the basis functions as a joint distribution problem, instead of specifying them manually. A mixture of ProMPs can be efficiently estimated by fitting a GMM to the datapoints $\bm{w}_m$, and using the linear transformation property of Gaussians to convert this mixture into a mixture at the trajectory level. Moreover, such representation can be extended to other basis functions, including Bernstein and Fourier basis functions, see Fig.~\ref{fig:ProMP2}-\emph{right}.

ProMP has been demonstrated in various robotic tasks requiring human-like motion capabilities such as playing the maracas and using a hockey stick~\cite{Paraschos13}, or for collaborative object handover and assistance in box assembly~\cite{Maeda16}. A Matlab code example \texttt{demo\_proMP01.m} can be found in~\cite{pbdlib}.

\section{Further challenges and conclusion}
This chapter presented various forms of superposition for time signals analysis and synthesis, by emphasizing the connections to Gaussian mixture models. The connections between these decomposition techniques are often underexploited, mainly due to the fact that these techniques were developed separately in various fields of research. The framework of mixture models provides a unified view that is inspirational to make links between these models. Such links also stimulate future developments and extensions. 

Future challenges include a better exploitation of the joint roles that mixture of experts (MoE) and product of experts (PoE) can offer in the treatment of time series and control policies~\cite{Pignat19}. While MoE can decompose a complex signal by superposing a set of simpler signals, PoE can fuse information by considering more elaborated forms of superposition (with full precision matrices instead of scalar weights). Often, either one or the other approach is considered in practice, but many applications would leverage the joint use of these two techniques. 

There are also many further challenges specific to each basis function categories presented in this chapter. For Gaussian mixture regression (GMR), a relevant extension is to include a Bayesian perspective to the approach. This can take the form of a model selection problem, such as an automatic estimation of the number of Gaussians and rank of the covariance matrices~\cite{Tanwani19IJRR}. This can also take the form of a more general Bayesian modeling perspective by considering the variations of the mixture model parameters (including means and covariances)~\cite{Pignat19}. Such extension brings new perspectives to GMR, by providing a representation that allows uncertainty quantification and multimodal conditional estimates to be considered. Other techniques like Gaussian processes also provide uncertainty quantification, but they are typically much slower. A Bayesian treatment of mixture model conditioning offers new perspectives for an efficient and robust treatment of wide-ranging data. Namely, models that can be trained with only few datapoints but that are rich enough to scale when more training data are available.

Another important challenge in GMR is to extend the techniques to more diverse forms of data. Such regression problem can be investigated from a geometrical perspective (e.g., by considering data lying on Riemannian manifolds~\cite{Jaquier17IROS}) or from a topological perspective (e.g., by considering relative distance space representations~\cite{Ivan13}). It can also be investigated from a structural perspective by exploiting tensor methods~\cite{Kolda09}. When data are organized in matrices or arrays of higher dimensions (tensors), classical regression methods first transform these data into vectors, therefore ignoring the underlying structure of the data and increasing the dimensionality of the problem. This flattening operation typically leads to overfitting when only few training data are available. Tensor representations instead exploit the intrinsic structure of multidimensional arrays. Mixtures of experts can be extended to tensorial representations for regression of tensor-valued data~\cite{Jaquier19}, which could potentially be employed to extend GMR representations to arrays of higher dimensions.

Regarding B\'ezier curves, even if the technique is well established, there is still room for further perspectives, in particular with the links to other techniques that such approach has to offer. For example, B\'ezier curves can be reframed as a model predictive control (MPC) problem~\cite{Egerstedt10,Berio17GI}, a widespread optimal control technique used to generate movements with the capability of anticipating future events. Formulating B\'ezier curves as a superposition of Bernstein polynomials also leaves space for probabilistic interpretations, including Bayesian treatments.

The consideration of Fourier series for the superposition of basis functions might be the approach with the widest range of possible developments. Indeed, the representation of continuous time signals in the frequency domain is omnipresent in many fields of research, and, as exemplified with ergodic control, there are many opportunities to exploit the Gaussian properties in mixture models by taking into account their dual representation in spatial and frequency domains.

With the specific application of ergodic control, the dimensionality issue requires further consideration. In the basic formulation, by keeping $K$ basis functions to encode time series composed of datapoints of dimension $D$, $K^D$ Fourier series components are required. Such formulation has the advantage of taking into account all possible correlations across dimensions, but it slows down the process when $D$ is large. A potential direction to cope with such scaling issue would be to rely on Gaussian mixture models (GMMs) with low-rank structures on the covariances~\cite{Tanwani19IJRR}, such as in mixtures of factor analyzers (MFA) or mixtures of probabilistic principal component analyzers (MPPCA)~\cite{Bouveyron13}. Such subspaces of reduced dimensionality could potentially be exploited to reduce the number of Fourier basis coefficients to be computed.

Finally, the probabilistic representation of movements primitives in the form of trajectory distributions also offers a wide range of new perspectives. Such models classically employ radial basis functions, but can be extended to a richer family of basis functions (including a combination of those). This was exemplified in the chapter with the use of Bernstein and Fourier bases to build probabilistic movement primitives, see Fig.~\ref{fig:ProMP2}-\emph{right}. More generally, links to kernel methods can be created by extension of this representation~\cite{Huang19}. Other extensions include the use of mixture models and associated Bayesian methods to encode the weights $\bm{w}_m$ in the subspace of reduced dimensionality.

\begin{acknowledgement}
I would like to thank Prof.\ Michael Liebling for his help in the development of the ergodic control formulation applied to Gaussian mixture models and for his recommendations on the preliminary version of this chapter.\\
The research leading to these results has received funding from the European Commission's Horizon 2020 Programme (H2020/2018-20) under the MEMMO Project (Memory of Motion, http://www.memmo-project.eu/), grant agreement 780684.
\end{acknowledgement}

\bibliographystyle{spbasic}
\bibliography{bib_timeSeries}

\end{document}